# Masked Conditional Neural Networks for Automatic Sound Events Recognition

Fady Medhat     David Chesmore     John Robinson

Department of Electronic Engineering
University of York, York
United Kingdom

{fady.medhat, david.chesmore, john.robinson}@york.ac.uk

*Abstract*— Deep neural network architectures designed for application domains other than sound, especially image recognition, may not optimally harness the time-frequency representation when adapted to the sound recognition problem. In this work, we explore the ConditionaL Neural Network (CLNN) and the Masked ConditionaL Neural Network (MCLNN)[1] for multi-dimensional temporal signal recognition. The CLNN considers the inter-frame relationship and the MCLNN enforces a systematic sparseness over the network's links to enable learning in frequency bands rather than bins allowing the network to be frequency shift invariant mimicking a filterbank. The mask also allows considering several combinations of features concurrently, which is usually handcrafted through exhaustive manual search. We applied the MCLNN to the environmental sound recognition problem using the ESC-10 and ESC-50 datasets. MCLNN achieved competitive performance, using 12% of the parameters and without augmentation, compared to state-of-the-art Convolutional Neural Networks.

## I. Introduction

Sound recognition is a wide research field that combines two broad areas of research; signal processing and pattern recognition. One of the very early attempts in sound recognition, especially speech, was in the work of Davis et al. [1] in 1952. In their work, they devised an analog circuitry for spoken digits' recognition. Their attempt marks a very early interest in the sound recognition problem. Over the years, the methods have evolved to involve not just speech, but music and environmental sound recognition as well. This interest was backed-up with the wide spread of related applications. For example, the usage of music sharing platforms or applications of automatic environmental sound recognition for surveillance [2, 3] especially when low lighting conditions hinders the ability of the video channel to capture useful information.

Handcrafting the features extracted from a signal, image or sound, has been widely investigated in research. The efforts invested aim to provide distinctive features that can enhance the recognition accuracy of the pattern recognition model. Recent attempts using deep neural networks have achieved breakthrough results [4] for image recognition. These deep models managed to abstract the features of a raw input signal over multiple neural network layers. The extracted features are further classified using a conventional classifier such as Random Forest [5] or Support Vector Machines (SVM) [6].

An attempt to use the deep neural network architectures for automatic feature extraction for sound was in the work of Hamel et al. [7]. In their work, they used three stacked Restricted Boltzmann Machines (RBM) [8] to form a Deep Belief Net (DBN) [9] architecture. They used the DBN for feature extraction from music clips. The extracted features were further classified using an SVM. They showed in their work the abstract representations captured by the RBM at each layer which consequently enhances the classification compared to using the raw time-frequency representation.

Deep architectures of Convolutional Neural Networks (CNN) [10] achieved remarkable results in image recognition [4]. Also, they got adapted to the sound recognition problem. For example, CNN was used in [11] for phoneme recognition in speech, where the CNN was used to extract the features, and the states' transitions were modeled using a Hidden Markov Model (HMM) [12].

Handcrafted features for sound are still superior in most contexts compared to employing neural networks as feature extractors of images, but the accuracy gap is getting narrower. The motivation behind using neural networks aims to eliminate the efforts invested in handcrafting the most efficient features for a sound signal.

Several neural based architectures have been proposed for the sound recognition problem, but usually, they get adapted to sound after they gain wide success in other applications especially image recognition. The adaptation of such models to sound may not harness its related properties in a time-frequency representation. For example, an RBM treats the temporal signal frames as static, isolated frames, ignoring the inter-frame relation. The CNN depends on weight sharing, which does not preserve the spatial locality of the learned features.

We discuss in this work, the ConditionaL Neural Network (CLNN) that is designed for multidimensional temporal signals. The Masked ConditionaL Neural Network (MCLNN) extends upon the CLNN by embedding a filterbank-like behavior within the network through enforcing a systematic sparseness over the network's weights. The filterbank-like pattern allows the

---

[1] Code: https://github.com/fadymedhat/MCLNN

This work is funded by the European Union's Seventh Framework Programme for research, technological development and demonstration under grant agreement no. 608014 (CAPACITIE).

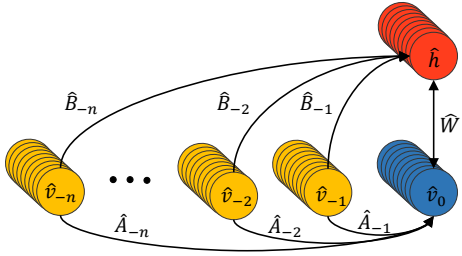

Fig. 1. Conditional Restricted Boltzmann Machine

network to exploit performance advantages of filterbanks used in signal analysis such as frequency shift-invariance. Additionally, the masking operation allows an automatic exploration of a range of feature combinations concurrently analogous to the manual features selection to be used for classification.

The models we discuss in this work have been considered in [13] for music genre classification with more emphasis on the influence of the data split (training set, validation set and testing set) on the reported accuracies in the literature. In this work, we evaluate the applicability of the models to sounds of a different nature i.e. environmental sounds.

## II. RELATED WORK

The Restricted Boltzmann Machine (RBM) [8] is a generative model that undergoes an unsupervised training. The RBM is formed of two layers of neurons, a visible and a hidden layer. The two layers are connected using bidirectional connections across them with the absence of connections between neurons of the same later. An RBM is trained using contrastive divergence [14] aiming to minimize the error between an input feature vector introduced to the network at the visible layer and the reconstructed version of the generated vector from the network.

We referred earlier that one of the drawbacks of applying an RBM to a temporal signal is ignoring the temporal dependencies between the signal's frames. The Conditional Restricted Boltzmann Machine (CRBM) introduced by Taylor et al. [15] extended the RBM [8] for temporal signals by adapting conditional links from the previous visible input to consider their influence on the network's current input. Fig. 1 shows a CRBM with the RBM represented by the visible $\hat{v}_0$ and hidden $\hat{h}$ layers with bidirectional connections $\widehat{W}$ going across them. The CRBM involves the conditional links from the previous visible input states to both the hidden $\hat{h}$ layer and the current visible input $\hat{v}_0$. The links between the previous input and the hiden layer are depicted by $(\hat{B}_{-1}, \hat{B}_{-2}, …, \hat{B}_{-n})$. The autoregressive links between previous visible input and the current one are depicted by $(\hat{A}_{-1}, \hat{A}_{-2}, …, \hat{A}_{-n})$. The CRBM was applied on a multi-channel temporal signal to model human motion through the joints movements. Mohamed et al. [16] applied the CRBM to the phoneme recognition task and they extended the CRBM in their work with the Interpolating CRBM (ICRBM) that considers the future frames in addition to the past ones. They showed in their work the outperformance of applying the ICRBM compared to the CRBM for the phoneme recognition task.

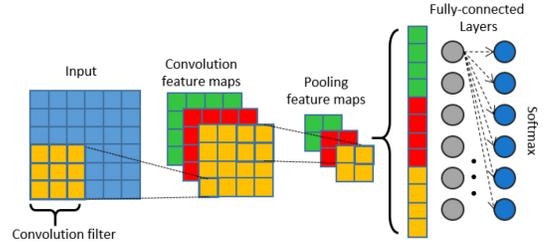

Fig. 2. Convolutional Neural Network

The Convolutional Neural Networks (CNN) [10] shown in Fig 2 depends on two main operations: Convolution and Pooling. In the convolution, the 2-dimensional input (an image) is scanned with several small sized weight matrices (filters), e.g. 5×5 in size. Each filter behaves as an edge detector on the input image. The output of the convolutional layer is a number of feature maps matching the number of filters used. The pooling stage involves decreasing the resolution of the generated feature maps, where mean or max pooling are usually utilized in this regard. Several of these two layers are interleaved to form deep architectures of neural networks, where the output of the final stage is flattened to a single feature vector or globally pooled [17] to be fed to a fully connected network for the classification decision. CNN assumes that a feature in a region of the input has a high probability of being located in other locations across the image. Accordingly, weight sharing is the fundamental concept of the CNN, which permitted applying neural networks to images of large sizes without having a dedicated weight for each pixel. The notion of weight sharing worked well for images, but it does not preserve the spatial locality of the learned features. Spatial locality of the features is an important consideration for spectrograms or time-frequency representations in general. The location of the learned features specifies the spectral component, where the same energy value may refer to different frequencies depending on the position at which it was detected. This induced attempts [18],[19],[20],[21], to tailor the CNN filters to the nature of the sound signal in a time-frequency representation in sound recognition for speech, environmental sounds and music.

## III. CONDITIONAL NEURAL NETWORKS

The ConditionaL Neural Network (CLNN) [13] is a discriminative model designed for temporal signals. The CLNN extends from the visible to hidden links proposed in the CRBM. Additionally, the CLNN considers the future frames in addition to the past ones as in the ICRBM. The CLNN takes into consideration the conditional influence, the frames in a window have on the window's middle frame, where the prediction of the central frame is conditioned on nearby frames on either side of it.

The input of a CLNN is a window of *d* frames, where *d* = 2*n* + 1. The order *n* specifies the frames to consider on either side of the window's middle frame (the 2 is to account for an equal number of frames in the past and the future and the 1 is for the window's middle frame). The hidden layer of a CLNN is an

$e$-dimensional vector of neurons. Accordingly, a CLNN generates a single vector of $e$-dimensions for each processed window of frames. The activation of a single node of a CLNN is given in (1)

$$y_{j,t} = f\left(b_j + \sum_{u=-n}^{n}\sum_{i=1}^{l} x_{i,u+t} W_{i,j,u}\right) \quad (1)$$

where $y_{j,t}$ is the activation of node $j$ of the hidden layer and the index $t$ refers to position of the frame within a segment (a chunk of frames with a minimum size equal to the window discussed later in detail), $f$ is the transfer function and $b_j$ is the bias of the neuron. $x_{i,u+t}$ is the $i^{th}$ feature of the $l$ dimensional input feature vector at index $u + t$ of the window. The frame's index $u$ in the window ranges from $-n$ up to $n$, where $t$ is the window's middle frame and in the same time, the index of the frame in the segment. $W_{i,j,u}$ is the weight between the $i^{th}$ feature of the input feature vector at index $u$ and $j^{th}$ hidden node. The vector form of the hidden layer activation is formulated in (2)

$$\hat{y}_t = f\left(\hat{b} + \sum_{u=-n}^{n} \hat{x}_{u+t} \cdot \widehat{W}_u\right) \quad (2)$$

where $\hat{y}_t$ is activation of the hidden layer for the frame at index $t$ of the input segment together with $n$ frames on either of its sides. $f$ is the transfer function and $\hat{b}$ is the bias vector. $\hat{x}_{u+t}$ is the input feature vector at index $u + t$, where $u$ is the index of the frame in the window and $t$ is the index of the window's middle frame in the segment. $\widehat{W}_u$ is the weight matrix at index $u$ within the window. Accordingly, for a window of $d$ frames a corresponding number of weight matrices are present in the weight tensor, where a vector-matrix multiplication operation is applied between the vector at index $u$ and its corresponding weight matrix at index $u$ in the weight tensor. The size of each weight matrix is [feature vector length $l$, hidden layer width $e$] and the count of matrices is equal to $2n+1$. The hidden layer's activation vector using a logistic transfer function is the conditional distribution of the prediction for window's middle frame conditioned on the order $n$ frames on either side. The relation is formulated as: $p(\hat{y}_t | \hat{x}_{-n+t}, ..., \hat{x}_{-1+t}, \hat{x}_t, \hat{x}_{1+t}, ..., \hat{x}_{n+t}) = \sigma(...)$, where $\sigma$ is a sigmoid function or the output softmax layer of the discriminative model.

Fig. 3 shows two stacked CLNN layers. Each CLNN layer possesses a weight tensor of size [*feature vector length l, hidden layer width e, window size d*] scanning the multidimensional in the temporal direction. The depth $d$ of the tensor matches the window size. Accordingly, for order $n=1$, the depth is 3, for $n=2$ the depth is 5 and for order $n$ the depth is $2n+1$ as the window size. At each layer, the number of frames decreases by $2n$ frames. Therefore, the size of the input segment follows (3) to account for the number of frames required at the input of a deep CLNN architecture.

$$q = (2n)m + k \quad , n, m \text{ and } k \geq 1 \quad (3)$$

where the $q$ is the number of frames in a segment, $m$ the number of layers and $k$ is the extra frames that should remain after the stacked CLNN layers. These $k$ frames can be flattened to a single feature vector or pooled across before introducing them to a fully connected network as shown in Fig. 3. For example, at $n = 4$, $m$

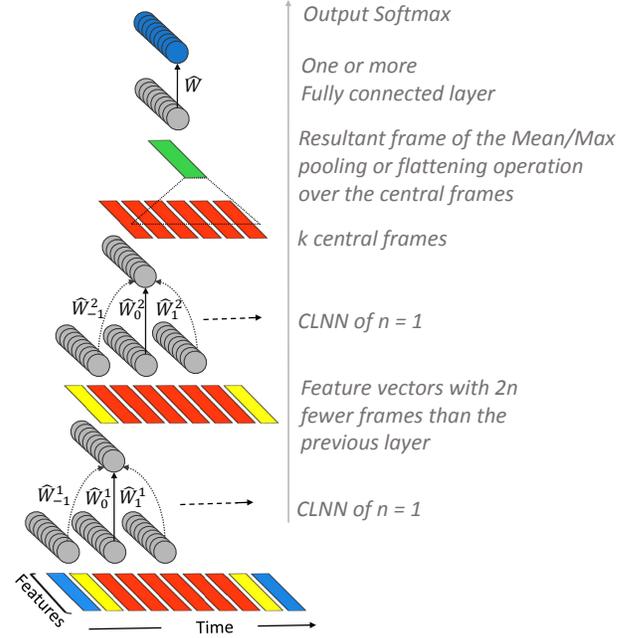

Fig. 3. A two layer CLNN model with n=1

= 3 (three CLNN layers) and $k$=5, the input at the first layer is $(2\times4) \times 3+5 = 29$. Therefore, the output size of the first layer is $29 - (2\times4) = 21$ frames. The output at the second layer is $21 - (2\times4) = 13$ frames. Finally, the output at the third layer is $13 - (2\times4) = 5$ frames. These remaining 5 frames are introduced to the fully connected layer after pooling or flattening to a single vector.

IV. MASKED CONDITIONAL NEURAL NETWORKS

Raw time-frequency representations such as spectrograms are used widely for signals analysis. They provide an insight of the changes in the energy across different frequency bins as the signal progresses through time. Due to their sensitivity to frequency fluctuations, a small shift in the energy of one frequency bin to a neighboring frequency bin changes the final spectrogram representations. A filterbank is a group of filters that allows subdividing the frequency bins, meanwhile aggregating the energy of each group of frequency bins into energy bands. Thus, suppressing the effect of the energy smearing across frequency bins in proximity to each other to provide a transformation that is frequency shift-invariant. For example, the Mel-scaled filterbank is formed of a group of filters having their center frequencies Mel-spaced from each other to follow the human auditory system perception of tones. Mel-scaled filterbanks are used in Mel-scaled transformations such as MFCC and Mel-Spectrogram both used widely by recognition systems as intermediate signal representations,

The Masked ConditionaL Neural Network (MCLNN) [13] extends upon the CLNN and stems from the filterbank by enforcing a systematic sparseness over the network's weights that follows a band-like pattern using a collection of ones and zeros as in Fig. 4. The masking embeds a filterbank-like behavior within the network, which induces the network to learn about frequency bands rather than bins. Learning in bands

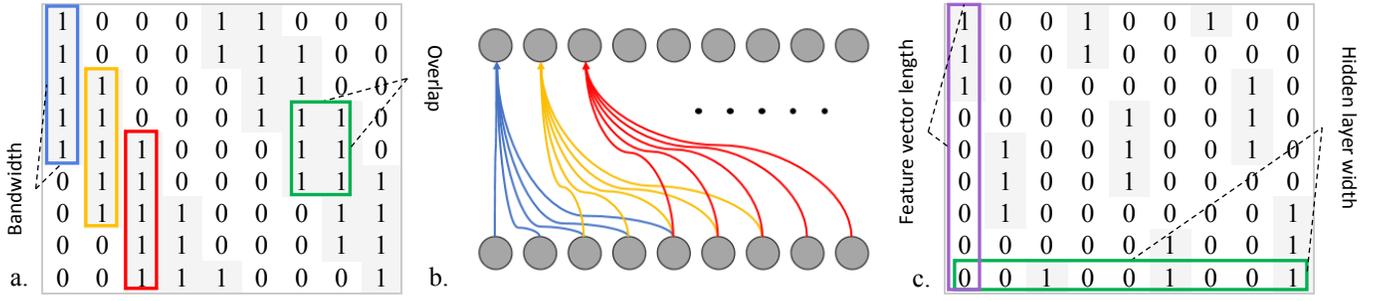

Fig. 4. Examples of the Mask patterns. a) A bandwidth of 5 with an overlap of 3, b) The allowed connections matching the mask in a. across the neurons of two layers, c) A bandwidth of 3 and an overlap of -1

prevents a hidden node from being distracted by learning about the whole input feature vector but instead allows a neuron to focus on a specific region in its field of observation, which permits distinctive features to dominate. The mask design depends on two tuneable hyper-parameters: the Bandwidth and the Overlap. The Bandwidth controls the number of successive 1's in a column, and the Overlap controls the superposition distance between one column and another. Fig. 4.a. shows a mask having a bandwidth of *5*, which refers to the enabled features in a vector and an overlap of *3*. Fig. 4.b shows the enabled network connections that map to the pattern in Fig. 4.a. The overlap can be assigned negative values as in Fig. 4.c, where an overlap of -1 refers to the non-overlapping distance between successive columns. The linear spacing of the binary pattern is formulated in (4)

$$lx = a + (g - 1)(l + (bw - ov)) \quad (4)$$

where the linear index *lx* is controlled by the bandwidth *bw*, the overlap *ov* and the feature vector length *l*. The values of *a* are within [0, *bw*-1] and *g* takes values in the interval $[1, \lceil (l \times e)/(l + (bw - ov)) \rceil ]$.

The handcrafting of the optimum features combination involves an exhaustive mix-and-match process aiming to find the right combination of features that can increase the recognition accuracy. The mask automates this process by embedding several shifted versions of the filterbank that allows combining different features in the same instance. For example, in Fig. 4.c, (the number of columns represents the hidden layer width) the first neuron in the hidden layer focuses on learning about the 1st three features of the input feature vector. Similarly, the fourth neuron (mapped to the 4th column in the mask) will learn about the first two features, and the 7th neuron will learn about one feature.

The masking operation is applied through an element-wise multiplication between the mask and each weight matrix present in the weight tensor following (5).

$$\hat{Z}_u = \widehat{W}_u \circ \widehat{M} \quad (5)$$

where $\widehat{W}_u$ is the matrix of size [*l, e*] at index *u* in a tensor of *d* weight matrices, $\widehat{M}$ is the masking pattern of size [*l, e*] and $\hat{Z}_u$ is the masked weight matrix to substitute the $\widehat{W}_u$ in (2)

Fig. 5 shows a single step of an MCLNN, where for a window of 2*n*+1 frames a weight tensor of a matching depth is processing the frames in the window. For each feature vector (frame) at a certain index, a corresponding matrix is processing it. The output of a single step of an MCLNN is a single representative vector representation. The highlighted cells in each matrix represent the masking pattern designed using the Bandwidth and the Overlap.

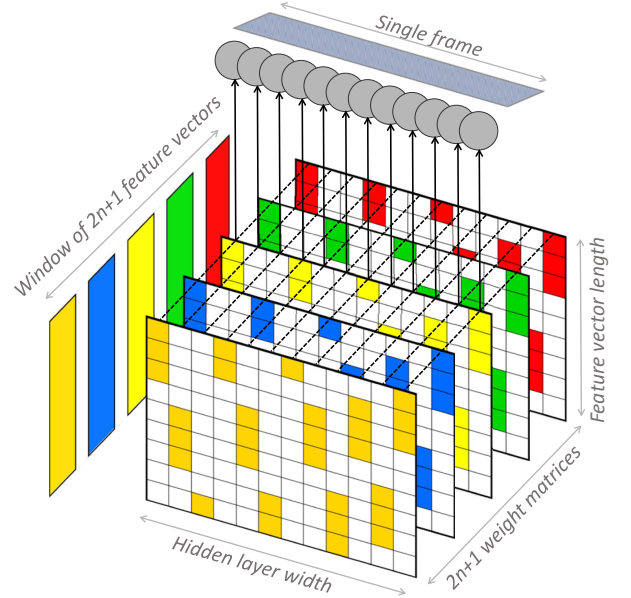

V. EXPERIMENTS

We used the ESC-10 [22] and the ESC-50 [22] datasets of environmental sounds to evaluate the performance of the MCLNN. Both datasets are released into 5-folds. The files are of 5 seconds each with files containing events shorter than 5 seconds padded with silence as described in [22]. As an initial preprocessing step, we trimmed the silence and cloned each file several times. We extracted 5 seconds from each of the cloned files. All files are resampled at 22050 Hz, followed by a 60 bin logarithmic Mel-scaled spectrogram transformation using an FFT window of 1024 and a 50% overlap with the Delta (1st derivative between frames across the temporal dimension). We concatenated the 60 bins and their delta column-wise resulting in a feature vector of 120 bin. We extracted segments of size *q* from each spectrogram following (3). All experiments followed the 5-fold cross-validation to unify reporting the accuracies by eliminating the influence of the data split. The training folds were standardized to a zero mean and unit variance feature-wise,

TABLE I  MCLNN HYPER-PARAMETERS

| Layer | Type | Nodes | Mask Bandwidth | Mask Overlap | Order $n$ |
|---|---|---|---|---|---|
| 1 | MCLNN | 300 | 20 | -5 | 15 |
| 2 | MCLNN | 200 | 5 | 3 | 15 |

and the training parameters were used to standardize the testing and validation sets.

We adopted two MCLNN layers with hyperparameters listed in Table I. The MCLNN layers are followed by a global pooling layer as studied in [17], but for the sound, it is a single dimensional mean global pooling layer. The pooling across the temporal dimension behaves as an aggregation operation, which enhances the accuracy as studied by Bergstra in [23]. Following the MCLNN layers are two densely connected layers of 100 neurons each before the final softmax output layer. We used Parametric Rectified Linear Units [24] for the activation functions. Dropout [25] as a regularizer. The model was trained to minimize the categorical cross-entropy between the predicted and the actual label of each segment of frames using ADAM [26]. Probability voting across the clip's frames was used for the category decision. We used FFmpeg [27] for the files cloning and LibROSA[28] for the signal transformation. For the model implementation, we used Theano [29] and Keras [30]

### A. ESC-10

The dataset is composed of 400 files for 10 environmental sound categories: Dog Bark, Rain, Sea Waves, Baby Cry, Clock Tick, Person Sneeze, Helicopter, Chainsaw, Rooster and Fire Cracking. The 400 sound files are equally distributed among the 10 classes with 40 clips per category.

For the ESC-10, We used $k = 1$, which leaves the window's middle frame remaining after the MCLNN layers to be fed to two fully connected layers. The human recognition accuracy for the dataset is 95.7%, and a baseline accuracy of 72.7% was achieved using Random Forest to classify MFCC frames in [22]. Table II lists the mean accuracies across a 5-fold cross validation on the ESC-10 dataset.

TABLE II  PERFORMANCE ON ESC-10 DATASET USING MCLNN COMPARED WITH OTHER ATTEMPTS IN THE LITERATURE

| Classifier and Features | Acc. % |
|---|---|
| **MCLNN + Mel-Spec. *without Augmentation* (this work)** | **83.0** |
| Piczak-CNN + Mel-Spec. *with Augmentation* [31] | 80.0 |
| **CLNN + Mel-Spec. *without Augmentation* (this work)** | **73.3** |
| Random Forest + MFCC *without Augmentation* [22] | 72.7 |

The work of Piczak [31] achieved 80% using a deep CNN architecture. The Piczak-CNN is formed of two convolutional and two pooling layers followed by two fully connect layers of 5000 neurons each resulting in a model containing over 25 million parameters. Piczak used 10 augmentation variants for each sound clip in the ESC-10 dataset. Augmentation involves introducing deformations to the sound files such as time delays and pitch shifting. Augmentation increases the dataset size, which consequently increases the generalization of the model and eventually the accuracy as studied by Salamon in [21]. We did not consider augmentation as it is not relevant in benchmarking the MCLNN performance against other models.

Dog Bark(DB), Rain (Ra), Sea Waves (SW), Baby Cry(BC), Clock Tick(CT), Person Sneeze(PS), Helicopter(He), Chainsaw(Ch), Rooster (Ro) and Fire Cracking (FC)

Fig. 6. Confusion matrix for the ESC-10 dataset.

On the other hand, the MCLNN outperformed Piczak-CNN accuracy using 3 million parameters achieving 83% without augmentation. To further ensure that MCLNN accuracy is not influenced by the intermediate representation, we adopted the spectrogram transformation used for the Piczak-CNN (60 bin Mel-spec. and delta).

To evaluate the influence of the mask absence in the CLNN, we used the exact architecture of the MCLNN. The CLNN achieved an accuracy of 73.3% compared to the 83% achieved by the MCLNN of the same architecture, which shows the effect of the masking operation due to the properties discussed earlier.

Fig. 6 shows the confusion across the ESC-10 dataset using the MCLNN. Clock Ticks is confused with Fire Cracking and Rain, which could be due to the short event duration. There is also apparent confusion between the Helicopter sound and the rain, which is accounted to the common low tonal components across the two categories.

### B. ESC-50

The dataset is formed of 2000 environmental sounds clips evenly distributed across the 50 classes of the following five broad categories:

*Animal sounds:* Hen, Cat, Frog, Cow, Pig, Rooster, Dog, Sheep, …etc.
*Natural soundscapes and water sounds e.g.* Sea waves, Rain, Thunderstorm, Crickets, Chirping birds, …etc.
*Human (non-speech) sounds e.g.* Snoring, Tooth brushing, Laughing, Footsteps, Coughing, Breathing, …etc.
*Domestic sounds e.g.* Glass breaking, Clock Ticking, Clock Alarm, Vacuum cleaner, Washing machine, …etc.
*Urban noise e.g.* Airplane, Train, Engine, Car horn, Siren, …etc.

The model we adopted for the ESC-50 is the same one we used for the ESC-10 dataset except for the order $n = 14$ and $k = 6$. The spectrogram transformation followed the same transformation employed by Piczak in [31] as described earlier.

TABLE III  Performance on ESC-50 dataset using MCLNN compared with other attempts in the literature

| Classifier and Features | Acc. % |
|---|---|
| Piczak-CNN + Mel-Spec. *with Augmentation* [31] | 64.50 |
| **MCLNN + Mel-Spec. *without Augmentation* (This Work)** | **61.75** |
| **CLNN + Mel-Spec. *without Augmentation* (This Work)** | **51.80** |
| Random Forest + MFCC *without Augmentation* [22] | 44.00 |

Table III lists the mean accuracies of a 5-fold cross-validation achieved on the ESC-50 dataset. MCLNN achieved 61.75% without augmentation and using 12% of the 25 million parameters employed in the Piczak-CNN. The CNN model used by Piczak in [31] is the same model applied to the ESC-10 dataset. Additionally, Piczak applied 4 augmentation variants for each sound clip in the ESC-50 dataset, which increases the accuracy as discussed earlier. The CLNN achieved 51.8% compared to the 61.75% achieved using an MCLNN, which supports similar findings regarding the masking influence on the results reported for the ESC-10.

## VI. Conclusion and Future Work

We have explored the possibility of applying the ConditionaL Neural Network (CLNN) designed for temporal signals, and its variant the Masked ConditionaL Neural Network (MCLNN) for environmental sound recognition. The CLNN is trained over a window of frames to preserve the inter-frames relation, and the MCLNN extends the CLNN through the use of a masking operation. The mask enforces a systematic sparseness over the network links, inducing the network to learn about frequency bands rather than bins. Learning in bands mimics the behavior of a filterbank allowing the network to be frequency shift-invariant. Additionally, the mask design allows the concurrent exploration of several features combinations analogous to handcrafting the optimum combination of features for a classification problem. We evaluated the MCLNN on the ESC-10 and the ESC-50 datasets of environmental sounds. The MCLNN achieved competitive accuracies compared to state-of-the-art Convolutional Neural Networks. MCLNN used 12% of the parameters used by CNN architectures of a similar depth. Additionally, MCLNN did not use augmentation adopted by the CNN attempts for the referenced datasets. Future work will consider deeper MCLNN architectures and different masking patterns. We will also consider applying the MCLNN to other multi-channel temporal signals other than spectrograms.